\documentclass{article}

\usepackage[final]{nips_2017}

\usepackage[utf8]{inputenc} 
\usepackage[T1]{fontenc}    
\usepackage{hyperref}       
\usepackage{url}            
\usepackage{booktabs}       
\usepackage{amsfonts}       
\usepackage{nicefrac}       
\usepackage{microtype}      

\usepackage{titlesec} 
\usepackage{bm} 
\newcommand{\vect}[1]{\boldsymbol{\mathbf{#1}}} 
\usepackage[super]{nth}
\usepackage{graphicx} 
\usepackage[symbol]{footmisc}

\renewcommand{\thefootnote}{\fnsymbol{footnote}}

\titleformat{\subsubsection}[runin]
  {\normalfont\normalsize\bfseries}{\thesubsubsection}{1em}{}

\usepackage{xcolor}
\usepackage{mypackage}

\newcommand{\npatients}{500 } 
\newcommand{\binsize}{100 } 

\usepackage{perpage} 
\MakePerPage{footnote} 
\usepackage{subcaption}
\usepackage{cleveref}

\title{Precision medicine as a control problem: Using simulation and deep reinforcement learning to discover adaptive, personalized multi-cytokine therapy for sepsis}

\makeatletter
\renewcommand*{\@fnsymbol}[1]{\ensuremath{\ifcase#1\or 1\or 2\or *\or
 \else\@ctrerr\fi}}
\makeatother

\author{
  Brenden K. Petersen\thanks{Lawrence Livermore National Laboratory, Livermore, CA 94550} \\
  \texttt{petersen33@llnl.gov} \\
  \And
  Jiachen Yang\footnotemark[1] \\
  \texttt{yjiachen@gmail.com} \\
  \And
  Will S. Grathwohl\footnotemark[1] \\
  \texttt{wgrathwohl@cs.toronto.edu}
  \And
  Chase Cockrell\thanks{University of Chicago, Chicago, IL 60637} \\
  \texttt{chase.cockrell@gmail.com}
  \And
  Claudio Santiago\footnotemark[1] \\
  \texttt{santiago10@llnl.gov}
  \And
   Gary An\footnotemark[2] \\
  \texttt{docgca@gmail.com}
  \And
  Daniel M. Faissol\footnotemark[1]\protect\phantom{\footnotesize 1}\textsuperscript{,}\thanks{Corresponding author}\\
  \texttt{faissol1@llnl.gov}
}

\begin{document}

\maketitle
\newcommand\blfootnote[1]{%
  \begingroup
  \renewcommand\thefootnote{}\footnote{#1}%
  \addtocounter{footnote}{-1}%
  \endgroup
}
\begin{abstract}
\blfootnote{This work was performed under the auspices of the U.S. Department of Energy by Lawrence Livermore National Laboratory under contract DE-AC52-07NA27344. Lawrence Livermore National Security, LLC. LLNL-JRNL-745693}Sepsis is a life-threatening condition affecting one million people per year in the US in which dysregulation of the body's own immune system causes damage to its tissues, resulting in a $28 - 50$\% mortality rate. Clinical trials for sepsis treatment over the last 20 years have failed to produce a single currently FDA approved drug treatment. In this study, we attempt to discover an effective cytokine mediation treatment strategy for sepsis using a previously developed agent-based model that simulates the innate immune response to infection: the Innate Immune Response agent-based model (IIRABM). Previous attempts at reducing mortality with multi-cytokine mediation using the IIRABM have failed to reduce mortality across all patient parameterizations and motivated us to investigate whether adaptive, personalized multi-cytokine mediation can control the trajectory of sepsis and lower patient mortality. We used the IIRABM to compute a treatment policy in which systemic patient measurements are used in a feedback loop to inform future treatment. Using deep reinforcement learning, we identified a policy that achieves 0\% mortality on the patient parameterization on which it was trained. More importantly, this policy also achieves 0.8\% mortality over 500 randomly selected patient parameterizations with baseline mortalities ranging from $1 - 99\%$ (with an average of 49\%) spanning the entire clinically plausible parameter space of the IIRABM. These results suggest that adaptive, personalized multi-cytokine mediation therapy could be a promising approach for treating sepsis. We hope that this work motivates researchers to consider such an approach as part of future clinical trials. To the best of our knowledge, this work is the first to consider adaptive, personalized multi-cytokine mediation therapy for sepsis, and is the first to exploit deep reinforcement learning on a biological simulation. 
\end{abstract}

\section*{Introduction}
Approximately one million people per year are diagnosed with sepsis in the United States, a life threatening condition in which the body's dysregulated immune response to infection or injury damages its own tissues and leads to organ failure \citep{singer2016}. The mortality rate for sepsis ranges from $28 - 50$\% \citep{wood2004}. While operational care process improvements in the last 20 years have led to reduction in mortality \citep{angus2015}, therapeutic options for sepsis are limited to variations of anti-microbial and physiological support dating back nearly a quarter century. There remains no FDA-approved, biologically targeted therapeutic for the treatment of sepsis. Attempts to discover biologically-targeted therapies for sepsis have thus far focused on manipulating a single mediator/cytokine, generally administered with either a single dose or over a very short course (< 72 hours) \citep{marshall2000clinical,opal2014next}. Cytokines are small proteins that regulate the immune system by having specific effects on the interactions and communications between cells. For example, cytokines can up-regulate inflammatory reactions, down-regulate inflammatory reactions, or induce cells to migrate (chemotaxis). Cytokine mediators are available for a number of cytokines, e.g. \citep{marshall2003such, reinhart2001anti, eichacker2002risk, zanotti2002cytokine}, and while none of these have been demonstrated to be clinically effective in treating sepsis, cytokine manipulation remains a primary target for anti-sepsis therapy development.  

Unfortunately, after more than 30,000 patients enrolled across various clinical trials costing hundreds of millions of dollars, none of the attempts to use cytokine mediators to cure sepsis have been successful \citep{marshall2000clinical,opal2014next}. These failures raise the question of whether sepsis can be controlled with cytokine mediation at all. Explanations for these failures that have been offered by researchers include inappropriate timing, duration, and/or dosing of the therapeutic agents as well as heterogeneity of the patient population, among other reasons. One possible explanation for these failures is that controlling sepsis with cytokine mediation requires an adaptive and personalized strategy that mediates multiple cytokines in coordination rather than a static and generic therapy targeting a single cytokine. Such a strategy is possible when a feedback loop is present such that the treatment strategy uses patient measurements during the course of treatment to inform future treatment.  An adaptive treatment strategy would result in mediation therapy changing over time for a single patient in response to the patient's specific progression. Indeed, the immune response is a complex, dynamic, and stochastic self-regulating system made up of heterogeneous entities that communicate with each other via a network of intermediate signaling proteins. Many engineered systems that are far simpler require complex and adaptive control strategies to maintain or recover equilibrium.
The focus of our study is to investigate whether the trajectory of clinical sepsis can be controlled with adaptive, personalized multi-cytokine mediation, by applying deep reinforcement learning to a surrogate dynamic computational model of sepsis.

For this investigation, we use a previously developed agent-based simulation model of systemic inflammation: the Innate Immune Response agent-based model (IIRABM) \cite{an2004} as a surrogate system for clinical sepsis. Simulation experiments with the IIRABM provide an explanation as to why single/limited cytokine perturbations at a single, or small number of, time points is unlikely to significantly improve the mortality rate of sepsis \citep{cockrell2017ga}. This result is consistent with the failures of previous static and non-personalized attempts at treating sepsis. Genetic algorithms have been used to explore and characterize multi-cytokine control strategies for the simulation that guide it from a persistent, non-recovering inflammatory state (functionally equivalent to the clinical states of systemic inflammatory response syndrome (SIRS) or sepsis) to a state of health \citep{cockrell2017ga}. While the calculated results show some generalizability, the authors recognized that more advanced strategies are needed to achieve the goal of adaptive personalized medicine in order to achieve lower simulated mortalities across a broader range of simulated patients. The genetic algorithm work inspired this investigation into adaptive, personalized multi-cytokine therapy.

To our knowledge, this work represents the first investigation into treating sepsis with adaptive and/or personalized cytokine mediation by exploiting systemic cytokine and immune cell measurements during the course of treatment. We hope that by demonstrating this strategy via simulation, we will help inform and direct research into ``minimally sufficient'' criteria in terms of sensor technology development, extent of interval sampling, biological targets for future drug development, and ultimately a clinical trial.  

Finally, this work presents the first application of deep reinforcement learning to control a simulation of a biological system. We believe the use of simulation together with reinforcement learning for discovering adaptive and personalized therapeutics can have broad applicability in the biomedical sciences, with the potential for large impacts.

\section*{Background}

\subsection{Agent-based models and their use as a sepsis simulation}

Agent-based models (ABMs) are a generic class of computational models well-suited to represent the complex, stochastic, and non-linear dynamics of biological systems such as the immune response \citep{an2009agent}. In an ABM, semi-autonomous software entities called \textit{agents} interact with each other and a virtual \textit{environment} by following a set of rules or operating principles. 

Often these local, individual interactions give rise to ``emergent'' or system-level dynamic behavior which otherwise cannot be easily inferred. ABMs can operate at multiple spatial and temporal scales, including those compatible with wet-lab and/or clinical measurements, thereby providing evidence for or against putative mechanisms of stochastic systems.

ABMs are a powerful alternative to differential equation-based models. While differential equation-based models excel in providing precise predictions for systems in which mechanisms are well-understood and uncertainty is low, ABMs facilitate exploring the behavior space for complex, stochastic systems in which mechanisms are poorly understood and uncertainty is high \citep{hunt2013}. ABMs can also be highly modularized \citep{petersen2014}, affording the ability to easily explore the space of new mechanisms and control strategies. ABMs have proven useful tools for biological applications, including immune system modeling \citep{baldazzi2006enhanced, bailey2007multi}, host-pathogen modeling \citep{bauer2009agent}, and cancerous tumor modeling \citep{wang2015simulating, zhang2007development}.

Specifically, we use the existing IIRABM as a surrogate system for the investigation of potential control strategies for a hospitalized patient with sepsis, a role previously proposed for ABMs \citep{an2017optimization}. We note that while the model does not contain a comprehensive list of all signaling mediators present in the human body, the most relevant cellular behaviors, namely those encompassing parallel and redundant pro- and anti-inflammatory processes, are represented. The named cytokines in this model are those that are most often associated with the available behavior rules in the current literature \citep{an2004}. The IIRABM simulates the human inflammatory signaling network response to injury. The model has been calibrated such that it reproduces the general clinical trajectories of sepsis \citep{an2004, cockrell2017abm}. The IIRABM operates by simulating multiple cell types and their interactions, including endothelial cells, macrophages, neutrophils, and helper T cells (T\textsubscript{h}0, T\textsubscript{h}1, and T\textsubscript{h}2 cells), as well as associated precursor or progenitor cells.

The  IIRABM is instantiated by setting five physiological parameters that specify the size and nature of the injury/infection as well as a measure of the host's resilience: initial injury size, microbial invasiveness, microbial toxigenesis, environmental toxicity, and host resilience. In previous work \citep{cockrell2017abm}, the authors performed a parameter sweep to determine the subset of all parameterizations that are considered clinically relevant. These are parameterizations that can lead to multiple outcomes: complete healing, death by infection, or death from immune dysregulation/sepsis.

Additionally, the  IIRABM has been used to simulate in silico clinical trials of mediator-directed therapies via the inhibition or augmentation of single cytokine synthesis pathways or limited hypothetical combinations of interventions \citep{an2004}. Those studies accurately reproduced unsuccessful clinical trials, as well as the non-efficacy of hypothetical interventions, but at the time could not provide a pathway toward discovering a successful intervention. 

\subsection{Reinforcement learning for controlling ABMs}
Reinforcement learning (RL) is a class of methods within machine learning for finding near-optimal solutions of a high-dimensional control problem that may not be analytically tractable. RL is a suitable candidate for controlling ABMs, which do not have analytic expressions for their global state dynamics and are not straightforward to approach using classical control methods. A traditional simulation of an ABM proceeds ``uninterrupted:'' given an initial state and starting parameters, agents behave semi-autonomously without external influence.  In the context of RL, however, external actions are applied throughout the course of the simulation in an attempt to guide the system toward a desired final state. The goal of RL is to learn the best action to take for each possible state during the simulation; it learns an adaptive \textit{policy} for maximizing a \textit{reward} function (e.g. a patient's health).

An RL problem is formulated as a Markov decision process, comprising five main elements: state space $\mathcal{S}$, action space $\mathcal{A}$, environment $\mathcal{E}$, reward function $r$, and an RL agent. (Note that the environment is distinct from the virtual environment in which IIRABM agents interact, and the RL agent is distinct from agents of the IIRABM.)

\begin{itemize}

\item{A \textit{state} $s \in \mathcal{S}$ is a complete specification of all quantities of the ABM at each time step.
While the ABM maintains its own internal state, the RL agent's observation $o = O(s)$ may be limited by a function $O(\cdot)$ that reduces the amount of accessible information. This is analogous to a clinician's limited observation of a patient's complete physiological state. To maintain a degree of clinical relevance, our RL agent will treat observations $o$ as the complete state of the system, and we shall use $s \in \mathcal{S}$ to denote the agent's state observation hereafter.
} 

\item{An \textit{action} $a \in \mathcal{A}$ is
is an external operation applied to the ABM at each time step, chosen by the RL agent rather than generated by the mechanistic rules of the ABM.}

\item{In this work, we recast the IIRABM as an OpenAI Gym environment \citep{brockman2016} to serve as the \textit{environment} $\mathcal{E}$, hereafter referred to as the sepsis environment. From the perspective of RL, $\mathcal{E}$ is a ``black box'' simulator: given an action $a_t$ chosen by the RL agent at time $t$, the sepsis environment updates its state from $s_t$ to $s_{t+1}$. A single trajectory through an environment---from the start of a simulation to its termination---is called an \textit{episode}.}

\item{The \textit{reward function} $r : \mathcal{S} \times \mathcal{A} \times \mathcal{S} \mapsto \mathbb{R}$ is a mapping from current state, the action taken, and the resulting state to a scalar signal at each time step. 
The \textit{return} is the discounted, cumulative future reward from time $t$ until the end of the episode, $R_t = \sum_{t'=t}^T \gamma^{t'-t} r_{t'}$, where $T$ is the terminal time step and $\gamma \in (0,1]$ is a \textit{discount factor} that determines the degree to which immediate rewards are favored over delayed rewards.}

\item{The \textit{RL agent} interacts with the environment by following a \textit{policy} $\pi : \mathcal{S} \mapsto \mathcal{P}(\mathcal{A})$, which, in the general case, maps a state to a distribution over actions.
In this work, we consider deterministic policies $\mu : \mathcal{S} \mapsto \mathcal{A}$, which can be learned with many fewer samples than stochastic policies \citep{silver2014deterministic}. An optimal policy $\mu^\star$ is one that maximizes the expected return.}

\end{itemize}

These five components interact in the following way: given a current observation $s_t$, the RL agent uses its policy to select an action $a_t = \mu(s_t)$ and applies it to the environment, which produces a new observation $s_{t+1}$ and an associated reward $r_t$.
The objective of an RL algorithm is to find an optimal or near-optimal policy by running repeated episodes through the environment and using the reward as a learning signal.

\subsection{Deep deterministic policy gradient}

RL algorithms fall under two general categories: value-function methods and policy search methods \citep{sutton1998reinforcement}. 
Among value-function methods, Q-learning is a widely-used algorithm for estimating an optimal action-value function $Q^{\star} : \Scal \times \Acal \mapsto \Rbb$, defined as $Q^{\star}(s, a) := \max_{\mu} \E \left[ R_t | s_t=s, a_t=a \right]$.
It gives the maximum expected return for executing action $a$ at state $s$ and following an optimal policy thereafter, and it induces a deterministic optimal policy $\mu^{\star} : \Scal \mapsto \Acal$ defined by $\mu^{\star}(s) := \argmax_{a \in \Acal} Q^{\star}(s,a)$.


In many systems, including the IIRABM, the state space $\mathcal{S}$ and/or action space $\mathcal{A}$ are both continuous and high-dimensional, rendering traditional tabular Q-learning approaches infeasible. Mnih et al. developed the \textit{deep Q-networks} (DQN) algorithm that adapted Q-learning to continuous and high-dimensional state spaces by utilizing an artificial neural network called the \textit{Q-network}, $Q(s,a;\theta)$, to parameterize the action-value function with neural network weights $\theta$. Training is executed using transitions $(s_t,a_t,r_t,s_{t+1})$ and stochastic gradient descent on a loss function $L(\theta) := \left(y_t - Q(s_t,a_t;\theta)\right)^2$, where $y_t := r_t + \gamma \max_{a'} Q(s_{t+1},a'; \theta)$. RL algorithms utilizing such networks are known as \textit{deep RL} (DRL) algorithms. Two main contributions led to the success of DQN: 1) the use of a \textit{target network} $Q'(s,a;\theta')$, which is a copy of the Q network used to evaluate $y_t$, but whose distinct weights $\theta'$ slowly update toward the learned weights $\theta$; and 2) the use of a \textit{replay buffer} containing a history of transitions, from which mini-batches are sampled for network training. DQN had great success in learning to play Atari video games, often achieving superhuman performance.

However, DQN cannot handle large, continuous action spaces. Lillicrap et al. extended DRL to continuous action spaces by utilizing an actor-critic framework that optimizes the performance $J(\theta^{\mu})$ of a policy $\mu(s;\theta^{\mu})$ \citep{lillicrap2015}. Their \textit{deep deterministic policy gradient} (DDPG) algorithm uses two main networks. An \textit{actor network} $\mu(s; \theta^\mu)$ outputs an action $a \in \Acal$ based on the current state $s$ and parameters $\theta^\mu$. A \textit{critic network} $Q(s, a; \theta^Q)$ evaluates the $Q$ value given the state-action pair and parameters $\theta^Q$. Actor network weights are updated using the deterministic policy gradient $\nabla_{\theta^{\mu}} J(\theta^{\mu}) = \mathbb{E}_{s \sim \mu} \left[ \nabla_{\theta^{\mu}} Q^{\mu}(s,\mu(s;\theta^{\mu})) \right]$ \citep{silver2014deterministic}. The critic is trained by gradient descent on the loss $L(\theta^Q) = (y_t - Q(s_t,a_t;\theta^Q))^2$, with $y_t = r_t + \gamma Q(s_{t+1},\mu(s_{t+1};\theta^{\mu}))$ known as the \textit{temporal difference} (TD) target. As with DQN, DDPG utilizes a replay buffer and target networks for the actor and critic. We employ DDPG to handle the continuous state and action spaces of the sepsis environment.


\section*{Related Work}

Here we review works that are related in some combination of methodology (i.e. DRL), type of environment (i.e. simulation, ABM), application domain (i.e. biology, sepsis), and use case (i.e. controlling a system to a desirable state).

Perhaps the most similar work is an application of traditional RL (i.e. not DRL) to control an ABM of tumor growth using radiotherapy. The authors employ tabular Q-learning \citep{jalalimanesh2017} to learn the optimal policy of radiotherapy, using an environment comprising an underlying ABM of vascular tumor growth. Both the action space (radiation intensity: choice of weak, normal, or intense) and observation space (tumor size: one of 200 bins) are one-dimensional and discretized for simplicity. Thus, DRL was not employed in this study since traditional tabular methods sufficed.


With respect to controlling sepsis, a recent study used DRL to discover retrospective policies for treating septic patients given a fixed clinical dataset comprising observations (e.g. patient vitals) and actions (i.e. administered medications) \citep{raghu2017}. Despite a similar goal to the present work, such retrospective studies using fixed datasets (a setting known as batch mode RL \citep{ernst2005}) are quite different in methodology, as there is no underlying simulation and thus no ability to query or explore the environment. Consequently, in contrast to our approach, only previously attempted therapeutics can be considered.  Also similar in goal but different in methodology is a recent approach using genetic algorithms to search for a non-adaptive control strategy using the same IIRABM as this work \citep{cockrell2017ga}.

There have been many recent advances in state-of-the-art DRL algorithms \cite{mnih2013, lillicrap2015} and variations of these algorithms \citep{vanhasselt2016, schaul2015, mnih2016, wang2015}. These approaches are typically benchmarked against standard RL environments (such as those curated by OpenAI Gym \citep{brockman2016}), including classic control problems, Atari 2600 games using the Arcade Learning Environment (ALE) \citep{bellemare2013}, board games (e.g. Go), and the MuJoCo physics engine \citep{todorov2012}. These benchmark environments are open-sourced and well-curated, facilitating performance comparisons across different algorithms. There are limited examples of DRL being applied to control a simulation (including an ABM). Notably, Li et al. and Casas use DRL for controlling the timing of traffic lights \citep{li2016, casas2017}. 

To our knowledge, our work represents the first application of DRL to control a simulation of a biological system.

\section*{Methods}
\subsection{Description of the IIRABM}

The IIRABM simulates the hospitalization of a patient diagnosed with sepsis. The virtual environment consists of a 101 $\times$ 101 grid representing a two-dimensional abstraction of the human endothelial-blood interface. Each grid point contains an endothelial cell agent and may contain additional leukocyte agents (a type of immune cell).  Each grid point also contains 14 scalar state variables: 12 cytokine concentrations, a measure of infection, and a measure of tissue damage. Agents (cells) contain additional state information, including age, activation status, and cytokine receptor concentrations.

The mechanisms of the IIRABM capture many cell-cell and cell-environment interactions related to the innate immune response and sepsis. The IIRABM includes mechanisms for infection spreading, tissue damage, chemotactic cell movement, leukocyte activation, leukocyte extravasation, respiratory burst, apoptosis, antibody administration, cytokine-receptor trafficking, and T cell differentiation. Details of these mechanisms are provided in \citep{an2004}.

Most relevant to the control problem is the mechanism by which cytokines interact with each other. Cytokine signaling ultimately governs the immune response; their dysregulation is a root issue underlying sepsis. Further, the simulated interventions we explore directly modulate these interactions. Thus, we describe this mechanism in detail. Under certain conditions, a cell agent updates the values of one or more cytokines at its grid point. In the absence of interventions, the general formula for updating the $i$\textsuperscript{th} cytokine is linear: $c_i \leftarrow \vect{\lambda} \cdot \vect{c},$ where $c_i$ is the cytokine being updated, $\vect{\lambda}$ is a vector of regulatory constants, and $\vect{c}$ is the vector of cytokine values at that grid point. The elements in $\vect{\lambda}$ represent the degree to which each cytokine up- or down-regulates cytokine $i$ via the cell's mediation. Additionally, direct cytokine secretion/elimination (the constant portion of the linear updates) is afforded by appending $\vect{c}$ with a dummy value of unity and appending $\vect{\lambda}$ with a value that represents the degree to which the cell directly secretes/eliminates cytokine $i$.

Cytokines are also altered via passive diffusion, spontaneous degradation, and external interventions. External interventions---or actions---map to putative cytokine-specific mediation therapies that either inhibit (down-regulate) or augment (up-regulate) cytokine signaling. Programmatically, an action $a \in \mathbb{R}$, specific to the the $i$\textsuperscript{th} cytokine, alters that cytokine's update rule by wrapping it in a function parameterized by $a$. Specifically, the update becomes $c_i \leftarrow f\left( \vect{\lambda} \cdot \vect{c} ; a \right),$ where $f$ is defined as:


$$
f(x; a) = \left\{
        \begin{array}{lll}
            10^a x & a \leq 0 & \textrm{(inhibition)} \\
            x + (10^a-1) & a \geq 0 & \textrm{(augmentation)} \\
        \end{array}
    \right.
$$

Thus, the wrapper function is multiplicative for negative (inhibitory) actions and additive for positive (augmentative) actions. Exponentiation is introduced so that the action space is symmetric and centered around $a = 0$. Note that when $a = 0$, the function becomes the identity function, i.e. it recovers the former update in which no intervention is applied. The control problem deals with dynamically choosing values of $a$ for each cytokine in a way that drives the simulated patient toward a healthy state.

A simulation begins with an applied bacterial infection. As it begins to spread, it quickly triggers cytokine signaling and an immune response. After 12 hr of simulated time, actions may be applied. This time delay reflects an approximation of the minimal time from onset of infection to presentation to a healthcare facility, identification of infection, and initiation of treatment. The simulation ultimately ends with one of two outcomes: complete healing or death. The health condition occurs when the system's total damage plus the total infection level (each expressed as a percentage of its maximum possible value) drops below 0.8\%. The 0.8\% offset is due to small, transient infection and damage caused by recurrent injuries to the host due to environmental conditions. Once intervention stops, to ensure that it provides a non-transient, lasting effect, a patient must proceed without intervention for at least 12 hr before the health condition is checked; this represents the fact that any putative intervention must cease. The death condition occurs when total damage exceeds 80\% (regardless of infection level) \citep{cockrell2017ga, an2017optimization}. 
Clinically, this threshold represents the ability of current medical technologies to keep patients alive (i.e. through organ support machines) in conditions that previously would have been lethal.



\subsection{Casting the IIRABM as an RL environment}



 
An episode begins 12 hr after the onset of the initial injury (i.e. there is a burn-in period of 12 hr from the perspective of the RL agent). By this time, the system has accumulated tissue damage and is trying to recover. An episode ends when the simulated patient completely heals or dies, or when the RL algorithm reaches 1,000 steps. The 1,000 step maximum introduces a third outcome during training, referred to as timeout.

Both RL and ABMs use the term ``step.'' To avoid ambiguity, hereafter we refer to a \textit{frame} as a single step of the ABM; we use \textit{step} to refer to a single step taken by the RL agent.

\subsubsection{Observation space.}

The IIRABM state exists over a discrete 101 $\times$ 101 grid. There are 21 real-valued state variables of interest at each grid point: concentrations for 12 cytokines, concentrations for 2 cytokine receptors, counts for 5 cell types, a measure of tissue damage, and a measure of infection. The grid is analogous to a digital image, in which each grid point represents a pixel and the state variables represent 21 channels (analogous to the 3 channels in a red-blue-green image).

When choosing the dimensionality of the observation space, there is an inherent trade-off between \textit{controllability} and \textit{clinical relevance}. At one extreme, we could provide the RL agent with an ``image'' representation; that is, spatially resolved measurements of all 21 dimensions. However, this observation space would be clinically unrealistic, as it would involve precisely resolving the locations of cells and local cytokine gradients in the body, which is infeasible given current measurement technology. More clinically relevant would be spatially aggregated values for each dimension, which are consistent with, for example, systemic cell and cytokine concentrations as measured from a blood sample.

Further, while spatial heterogeneity is important for the IIRABM state to evolve during the course of a simulation, we posit that resolving these spatial differences would not be critical for the control task. Specifically, we hypothesize that the spatially aggregated state space (each state variable summed across all grid points) provides sufficient information for learning a near-optimal policy. Thus, we reduced the putative observation space from $\mathbb{R}^{101 \times 101 \times 21}$ to $\mathbb{R}^{21}$.

Since the 21 input dimensions have different ranges and units and often exist on very different scales, we transformed each dimension by subtracting an offset and dividing the difference by a scaling factor. To select offsets and scaling factors, we first ran episodes without intervention and recorded empirical minimum and maximum values for each dimension. The offsets were calculated as the half the sum of the minimum and maximum value; the scaling factors were calculated as half the difference between the minimum and maximum value. The resulting transformed input dimensions thus tend to fall within $[-1, 1]$, though in practice (when actions are taken), the transformed inputs can fall outside this interval. We found that more automated methods of transforming inputs, e.g. normalizing inputs via batch normalization (as in \citep{lillicrap2015}) or other online normalization methods (e.g. whitening inputs via the mean and standard deviation of the current replay buffer), resulted in instability, perhaps due to shifts in the data distribution used to determine the transformation.

\subsubsection{Action space.}


As described in Methods, the action taken by the RL agent for cytokine $i$ corresponds to choosing the parameter $a_i$ that determines the amount of inhibition or augmentation by the update function $f(\mathbf{\lambda} \cdot \mathbf{c}; a_i)$.
We chose the joint action space for action vectors $\abf = (a_1, \dotsc, a_{12})$ to be the set $[-1,1]^{12} \subset \Rbb^{12}$. Clinically, the choice of continuous actions reflects the ability of multiple therapeutics to be administered independently and with precise control via intravenous infusions.

\subsubsection{Reward function.}




A fully healed patient receives a positive terminal reward (chosen as +250). A patient who dies receives a negative terminal reward (chosen as -250). If the episode terminates early due to the 1,000-step limit, there is no terminal reward.

Since the environment only naturally provides a terminal reward but episodes may be long, we sought an intermediate reward signal to facilitate learning. Specifically, we exploited potential-based reward shaping, a technique to guide the learning process in a way guaranteed not to change the optimal policy \citep{ng1999}. For this technique, a reward shaping signal $F(s_t, s_{t+1}) = \gamma\phi(s_{t+1}) - \phi(s_t)$ is added to the baseline reward signal, where $\phi : S \mapsto \mathbb{R}$ is called the \textit{potential function} and is only a function of state. Reward shaping functions of this form are proven to be \textit{policy invariant}; that is, an optimal policy when learning using the baseline plus potential-based reward signal will also be optimal when learning with only the baseline reward signal \citep{ng1999}.

Under the proof of policy invariance, the value of $\gamma$ used in $F$ is the same as that used in the TD target. However, there are practical disadvantages to using $\gamma < 1$ in the formula for $F$, depending on the sign and magnitude of the potential function \citep{grzes2009}; namely, if $\phi(s) \leq 0$ for all $s \in \mathcal{S}$ (as in our case), the value of $F$ can be positive even when $\phi(s_{t+1}) < \phi(s_t)$ (i.e. when there is a decrease in potential). Only when the fold-decrease in potential is large enough ($\frac{\phi(s_{t+1})}{\phi(s_t)} > \gamma^{-1}$) will the reward signal actually be penalized. This might preclude the RL algorithm from finding the optimal policy if changes to the potential function are small. To avoid this issue, we used $\gamma = 1$ in the reward shaping signal but preserved $\gamma = 0.99$ for the TD target.

A good target for the potential function is an estimate of the value of the state. Specifically, if $\phi$ is equal to the optimal value function in an RL formulation \textit{without} potential-based rewards, then the optimal value function will be identically equal to zero in the corresponding formulation \textit{with} potential-based rewards. A value function of zero greatly facilitates learning. Since termination conditions are based on total system damage, we defined the potential function as the negative of this value: $\phi(s) = -\textrm{damage}(s)$.

We added an $L_1$ penalty to the reward function for actions taken. This promotes conservative actions (ones that use less therapeutic), which is clinically relevant as real drugs incur additional side effects that are not captured in the model. 


Thus, the final reward function is:
$$
r(s_t, a_t, s_{t+1}) = \left\{
        \begin{array}{ll}
            -250 & \textrm{death} \\
            +250 & \textrm{health} \\
            \beta \left( \phi(s_{t+1}) - \phi(s_t) \right) - \lambda \Vert a_t \Vert_1 & \textrm{otherwise}, \\
        \end{array}
    \right.
$$

where $\beta = 100$ and $\lambda = 1$. Recall that the episode ends whenever health or death states are reached.


\subsection{Network architecture and hyperparameter selection}

We used the same network architecture and hyperparameters as \citep{lillicrap2015} with several exceptions. Briefly, both the actor and critic networks included two hidden layers (of 400 and 300 nodes, respectively) and rectified linear activation functions. For the critic network, a hyperbolic tangent activation was added to the output layer. For the critic network, actions were added at the second hidden layer. Differences from \citep{lillicrap2015} include that the inputs to each network were not batch normalized, which we found to result in instability as the data distribution of the experience replay buffer shifted during training. Further, batch normalization was not applied to hidden layers. For exploration, we added uncorrelated Gaussian noise with a standard deviation of 0.1 independently to each dimension of the selected action vector. This standard deviation was decreased 10-fold every 1,000 episodes.

Given a fixed reward value for the health/death outcome of $\pm 250$, the reward function has two hyperparameters: $\beta$ and $\lambda$. To ensure that the overall goal of reaching health/death remains the dominant driver of the reward function, we sought the contributions of potential-based reward shaping and penalizing actions to each be roughly two orders of magnitude less than the terminal rewards. We found that $\beta = 100$ and $\lambda = 1$ typically resulted in $\mathcal{O}(1)$ values for the potential-based term and action penalty term, respectively.

\subsection{Software and computation}

The IIRABM was implemented in C++ as in \citep{cockrell2017ga, an2017optimization} to maximize performance. We exposed it to Python using the Boost C++ libraries \citep{abrahams2003building} and recast it as an OpenAI Gym environment \citep{brockman2016}. We implemented DDPG in Python, leveraging TensorFlow \citep{abadi2016tensorflow} and Gym \citep{brockman2016} packages. Training was performed on an NVIDIA Titan X GPU.


\section*{Results and Discussion}

\subsection{Unique challenges of the IIRABM as an RL environment}

In principle, RL can be applied to a wide range of curated environments. However, the sepsis environment poses several challenges compared to most ALE environments. First, the sepsis environment is inherently and deeply stochastic, mimicking the heterogeneity of the clinical setting. In contrast, Atari games are deterministic \citep{hausknecht2015impact}. In fact, given a fixed parameter initialization for the IIRABM and only varying the pseudo-random number generator seed, repeated simulation runs (with no intervention) produce very different state trajectories and even final outcomes (life or death). In fact, a thorough characterization of the IIRABM \citep{cockrell2017abm} shows that the vast majority of the state space falls within a ``stochastic zone'' in which both outcomes (life and death) remain possible for a large portion of the episode.  The dynamics of the system are deterministic only for a fixed random seed \textit{and} a fixed policy. Therefore, because the policy is not fixed during training (due to exploration and the neural network parameters changing), a small deviation in actions taken has the same effect as reseeding the pseudo-random number generator, even when the actual random seed is fixed.

Stochasticity leads to a substantially more challenging RL problem, in part because it causes the ``signal'' used for learning to be very noisy. That is, under the same conditions, a given action is sometimes ``good'' and sometimes ``bad.'' This problem is exacerbated by the fact that episodes are relatively long (up to 4,603 steps, or 90 days of simulated time) and the IIRABM provides only one natural reward: a binary life or death outcome at the end of the episode. This results in a very difficult credit assignment problem, i.e. it is difficult to determine which of the many actions taken in a single episode are responsible for the observed outcome. 

In other applications, sparse rewards are often overcome by seeding the DRL agent with demonstrations of ``good'' episodes as discovered by, for instance, human players. However, in our case, the clinical problem of sepsis and its representation in the sepsis environment have not been solved by humans, and therefore such techniques are not applicable. Indeed, whether the sepsis environment can be completely solved (i.e. 0\% mortality for all plausible patient parameterizations) is open question.

Another additional challenge presented by the IIRABM is that the environment is computationally expensive. We denote the computational cost (in wall clock time) for running one step of an RL environment as $t\textsubscript{env}$ and the computational cost of performing a single training step as $t\textsubscript{train}$. Using the same algorithm, network architecture, and GPU machine as described in Methods, the $t\textsubscript{env}$:$t\textsubscript{train}$ ratio is roughly 1:100 for a typical ALE environment. In stark contrast, this ratio becomes 3:1 for the sepsis environment. Since the network architectures are the same, this amounts to a roughly 300-fold increase in $t\textsubscript{env}$ for the sepsis environment. This cost precludes exploiting certain methods like frame skipping, a technique commonly used for ALE environments that exploits the low $t\textsubscript{env}$:$t\textsubscript{train}$ ratio. By only providing the RL agent with every $n$\textsuperscript{th} and repeating the chosen action $n$ times (where $n$ is typically 3 or 4), frame skipping allows for completing more episodes for the same number of training steps (at the cost of lower temporal resolution of choosing actions). However, in the case where the $t\textsubscript{env}$:$t\textsubscript{train}$ ratio exceeds 1:$n$, frame skipping actually results in fewer completed episodes per wall clock time.

\subsection{Training}

We trained the RL agent on a single patient parameterization with a baseline mortality rate (mortality rate without intervention) of 46\%. Each episode used a different random seed. We ceased training at 3,500 episodes (\textasciitilde 9.2 million frames), by which time the reward signal began to plateau and many episodes resulted in healing during training. Since we are primarily interested in controlling patient outcome, we saved all policies that resulted in a multiple of 20 consecutively healed patients during training.

\begin{figure}
  \centering
    \includegraphics[width=0.75\textwidth]{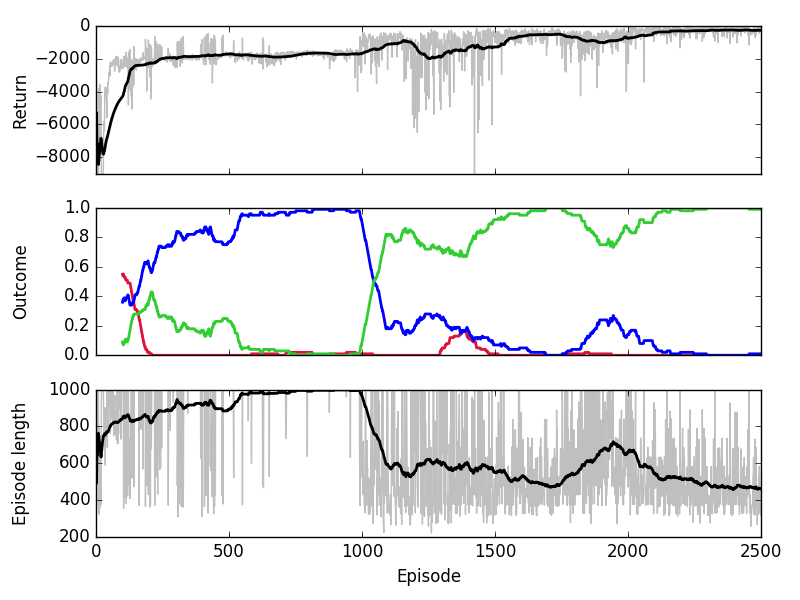}
  \caption{Training performance. Top: Return per episode. Middle: Moving average over 100 episodes of the rates of patient outcome (death: red, timeout: blue, health: green). Bottom: episode length (in steps). For the top and bottom plots, gray represents the actual value each episode; black represents a moving average over 100 episodes.}
  \label{fig:log}
\end{figure}

Figure \ref{fig:log} illustrates the return, outcome distribution, and episode length as a function of training episode. Taken together, these signals suggest that the RL agent progresses through several qualitatively different stages of learning. Early on, some patients heal during training, as the policy begins centered around zero-valued actions (recall that the patient heals 46\% of the time without intervention). The return then reaches an early peak around episode 125, despite a poor healing rate and long episodes. As the RL agent continues to learn, episode lengths increase up to the 1,000 step maximum as both the healing and death rates drop to nearly zero. Around episode 1,000, the RL agent  begins healing patients quickly and episode length correspondingly declines to roughly 500 steps. For the remainder of the training episodes, the policy maintains a high healing rate. Thus, it appears that the RL agent first learned to stabilize the patient (leading to long episodes) despite not healing patients. It then learned to heal these patients and continued to improve by healing them faster.

\subsection{Policy evaluation and generalizability}

The reward signal is a construct engineered to help guide the policy toward healing patients; the clinically relevant measure of interest is not return but outcome (life or death). Since the simulation is stochastic, any given initialization (with or without intervention) results in a distribution of outcomes. As such, we evaluate a learned policy by its resulting mortality rate rather than its average return. Specifically, a policy proceeded until the strict healing threshold (total damage plus total infection less than 0.02\%) was met or 1,000 time steps elapsed. At this point, interventions stopped and the patient proceeded for at least 12 hr without intervention before determining the final outcome. The learned policy resulted in 0\% patient mortality during test time for the patient parameterization on which DDPG was trained.

We analyzed the generalizability of the learned policy by testing it over a set of 500 different patient parameterizations with baseline mortality rates of $1 - 99\%$. The patient parameterizations included in this study span the entire space of plausible parameterizations for the IIRABM as determined in \citep{cockrell2017abm}.
To ensure our test set of patient parameterizations also had a good spread of baseline mortality rates, we combined five subsets of patient parameterizations, each of which comprised \binsize randomly selected patient parameterizations with baseline mortality rates in one of the following intervals: $(1\%, 20\%), [20\%, 40\%), [40\%, 60\%), [60\%, 80\%), [80\%, 99\%).$ For each patient parameterization, post-intervention mortality rate was calculated over 50 episodes with different random seeds. 

\begin{figure}
  \centering
    \includegraphics[width=0.75\textwidth]{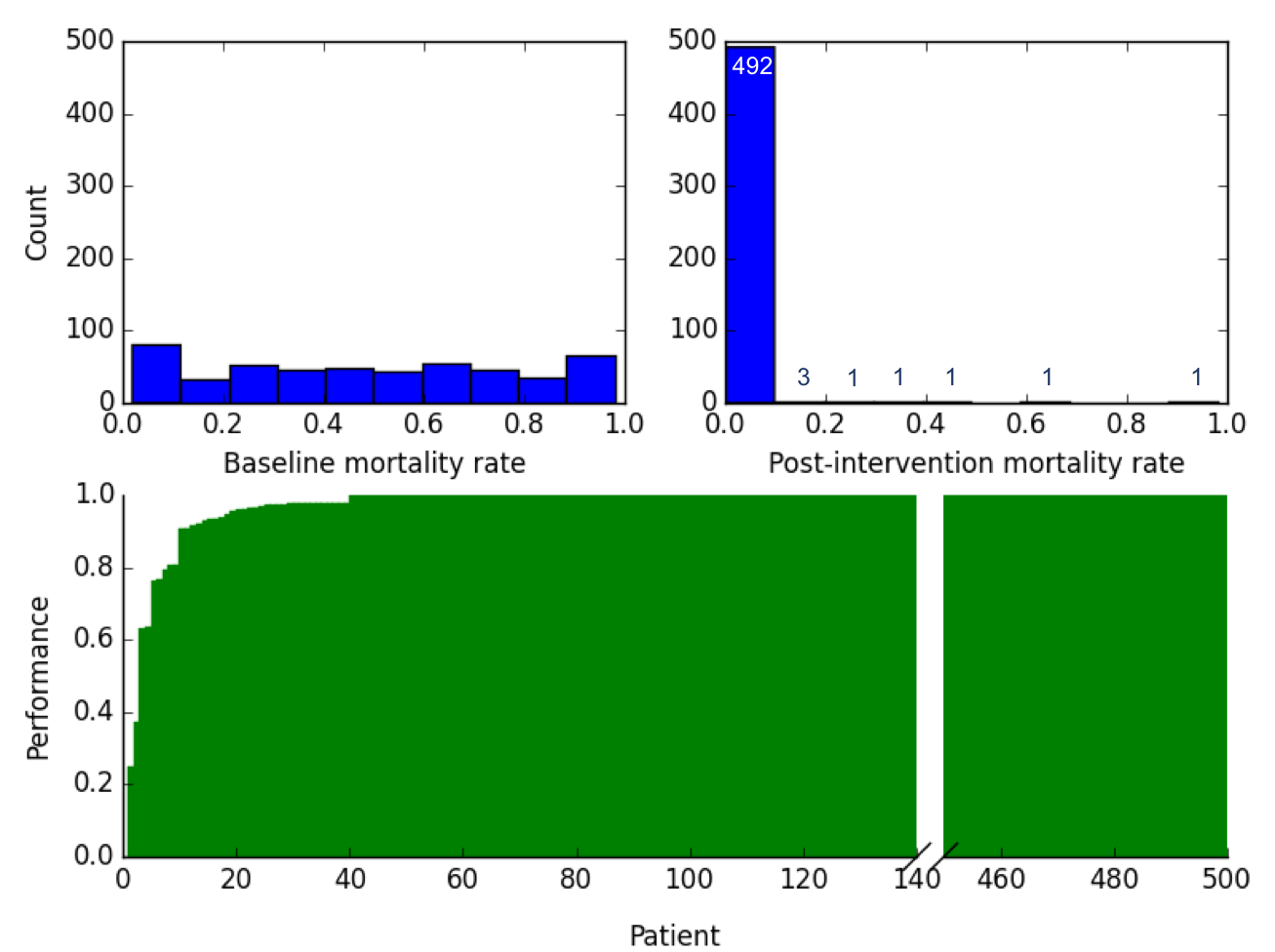}
  \caption{Top-left: Histogram of baseline mortality rates for \npatients test patient parameterizations (N = 50 instances per test patient parameterization). Top-right: Histogram of post-intervention mortality rates for \npatients test patient parameterizations. Bottom: Performance of the learned policy on each of the \npatients test patient parameterizations, calculated as $(\textrm{baseline mortality} - \textrm{post-intervention mortality})/\max(\textrm{baseline mortality, post-intervention mortality})$, sorted by increasing performance.}
  \label{fig:mortality}
\end{figure}

The mortality rates using the learned policy are illustrated as a histogram in Figure \ref{fig:mortality}. The overall mortality rate (across all \npatients patient parameterizations) changed from 46.0\% without intervention to 0.8\% under the learned policy. Additionally, under the learned policy, 460 of the \npatients patient parameterizations (92\%) have a mortality rate of 0\%, 39 patient parameterizations (7.8\%) exhibited reduced mortality rate (with an average reduction of 87\%), and 1 patient parameterization resulted in no change in mortality rate compared to baseline. 

For each test patient parameterization, we assessed the performance of the policy relative to the case with no intervention. Performance was calculated as the difference between baseline mortality and post-intervention mortality, normalized by the larger of the two mortality rates. Thus, positive values can be interpreted as the fraction of patients who were healed by the policy that otherwise would have died, while negative values indicate the fraction of patients who died from the policy that otherwise would have healed. Figure \ref{fig:mortality} (bottom) illustrates this performance metric for each patient parameterization. Since the learned policy did not cause an increase in mortality for any patient parameterization, the figure includes no negative values. These results suggest that despite being trained on a single patient parameterization, the learned policy generalizes well, as it is robust to changes in underlying parameterization.

\subsection{Policy characterization}

\begin{figure}
\begin{subfigure}{0.33\textwidth}
\centering
\includegraphics[width=\linewidth]{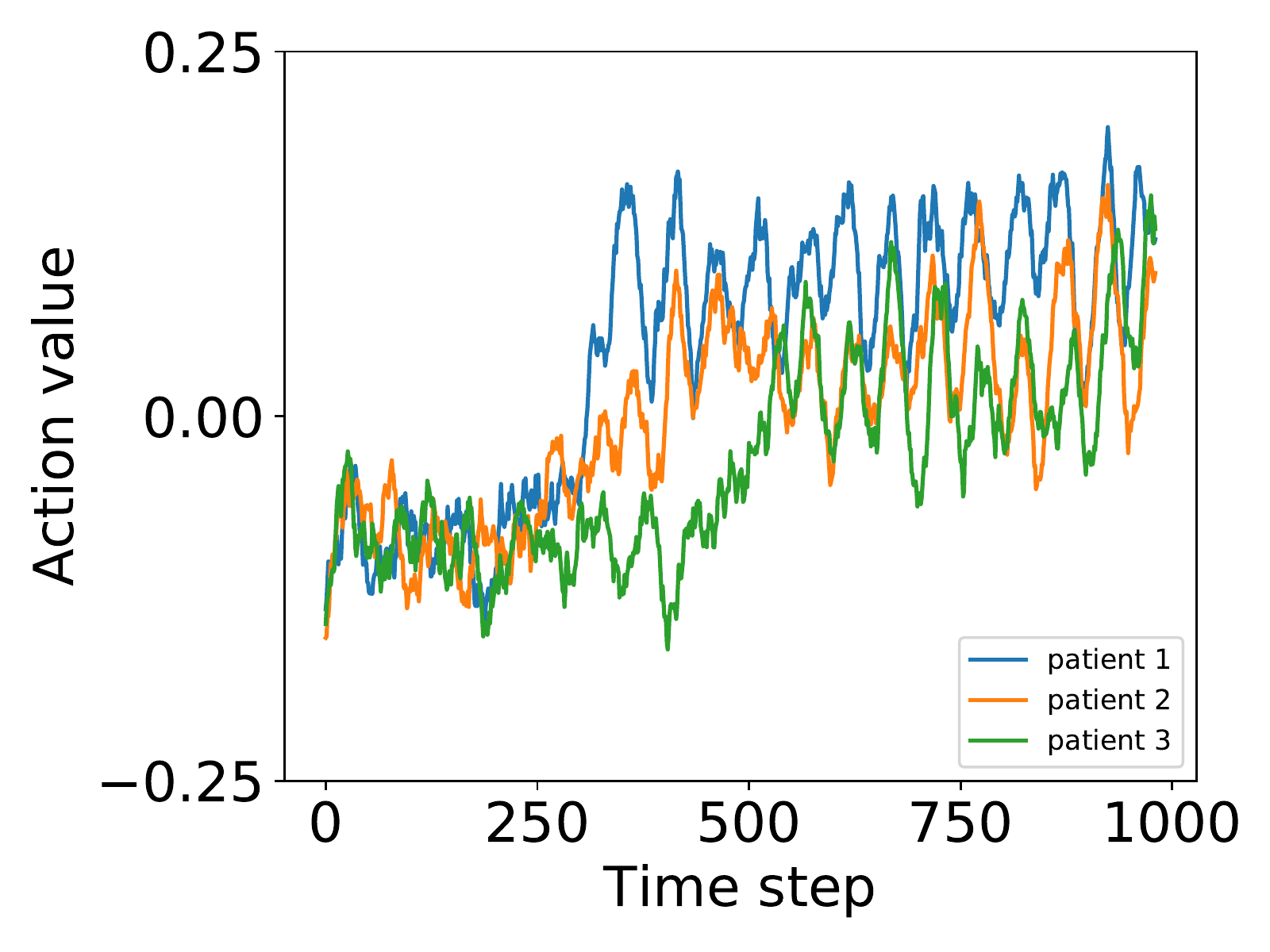}
\caption{PAF}
\label{fig:actions-1}
\end{subfigure}
\begin{subfigure}{0.33\textwidth}
\centering
\includegraphics[width=\linewidth]{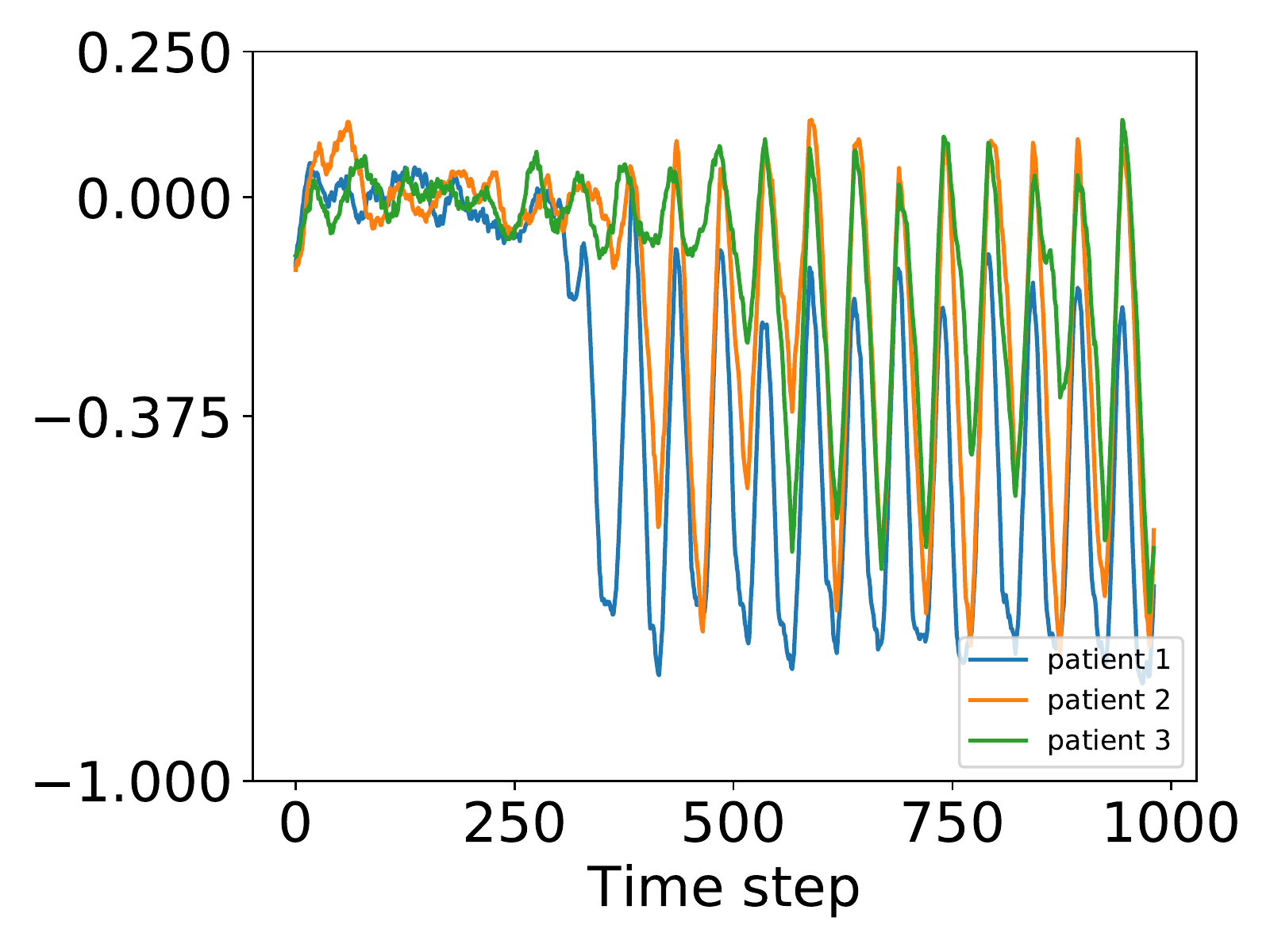}
\caption{IL-1}
\label{fig:actions-2}
\end{subfigure}
\begin{subfigure}{0.33\textwidth}
\centering
\includegraphics[width=\linewidth]{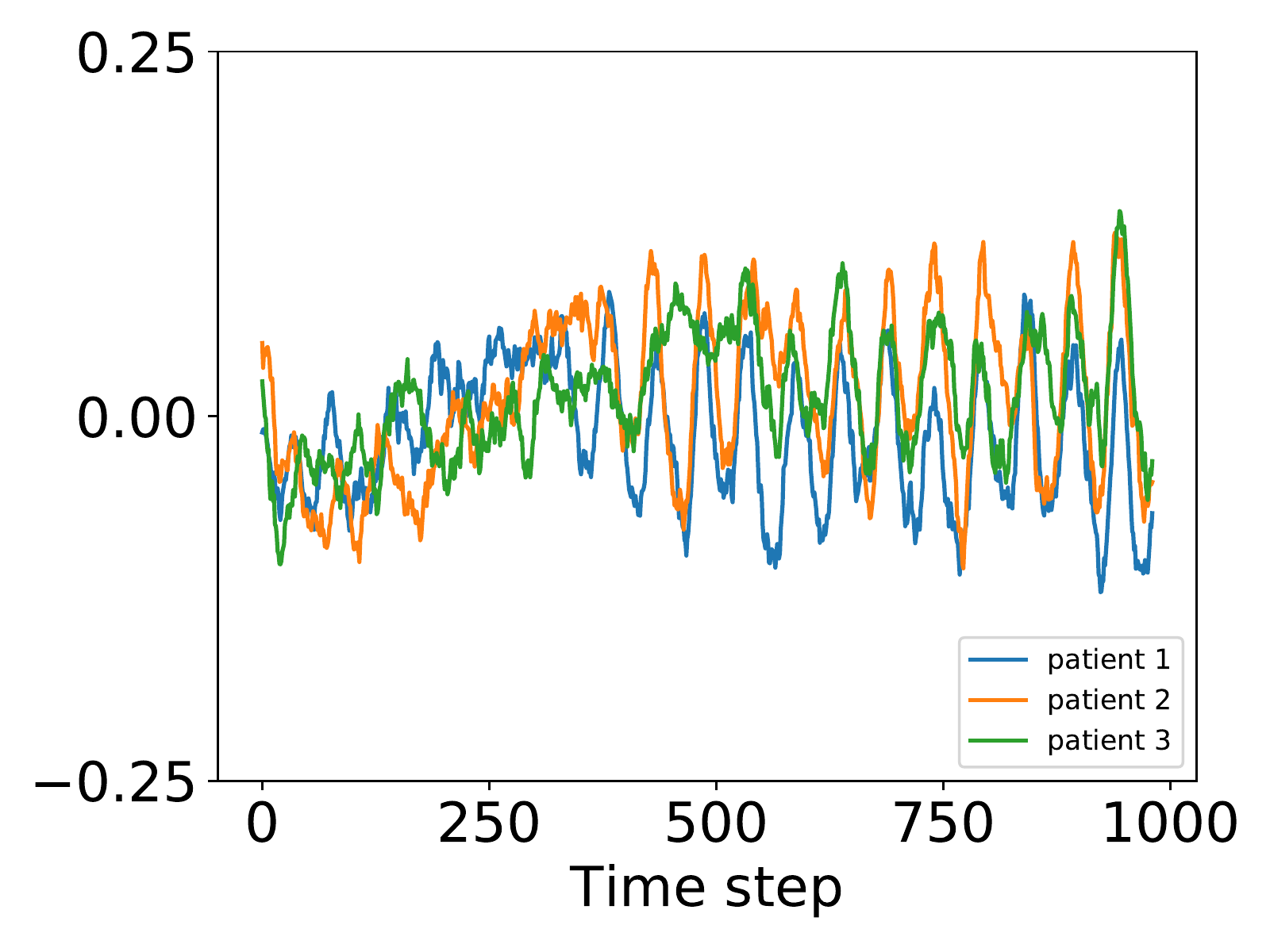}
\caption{IFN$\gamma$}
\label{fig:actions-3}
\end{subfigure}
\begin{subfigure}{0.33\textwidth}
\centering
\includegraphics[width=\linewidth]{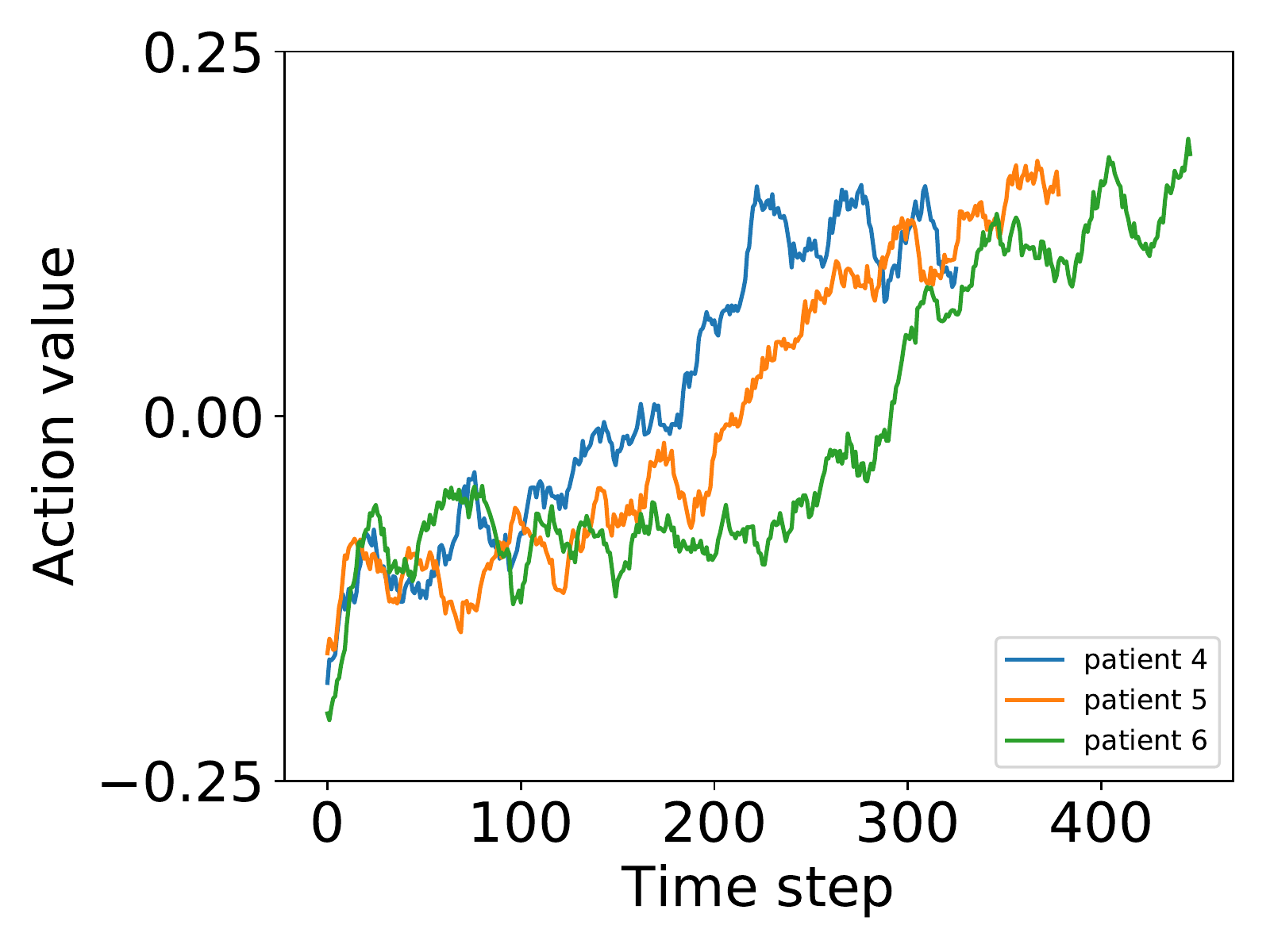}
\caption{PAF}
\label{fig:actions-4}
\end{subfigure}
\begin{subfigure}{0.33\textwidth}
\centering
\includegraphics[width=\linewidth]{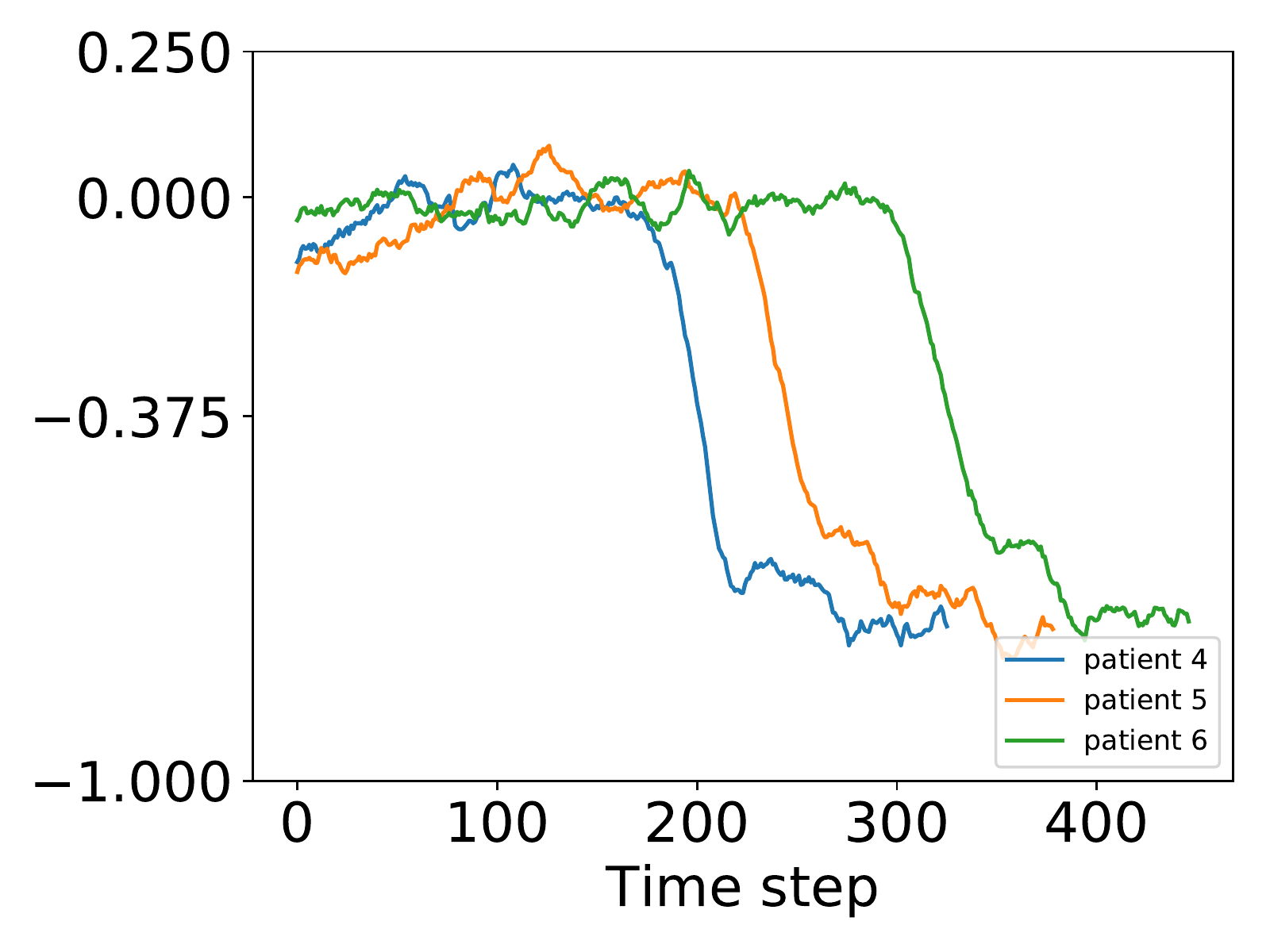}
\caption{IL-1}
\label{fig:actions-5}
\end{subfigure}
\begin{subfigure}{0.33\textwidth}
\centering
\includegraphics[width=\linewidth]{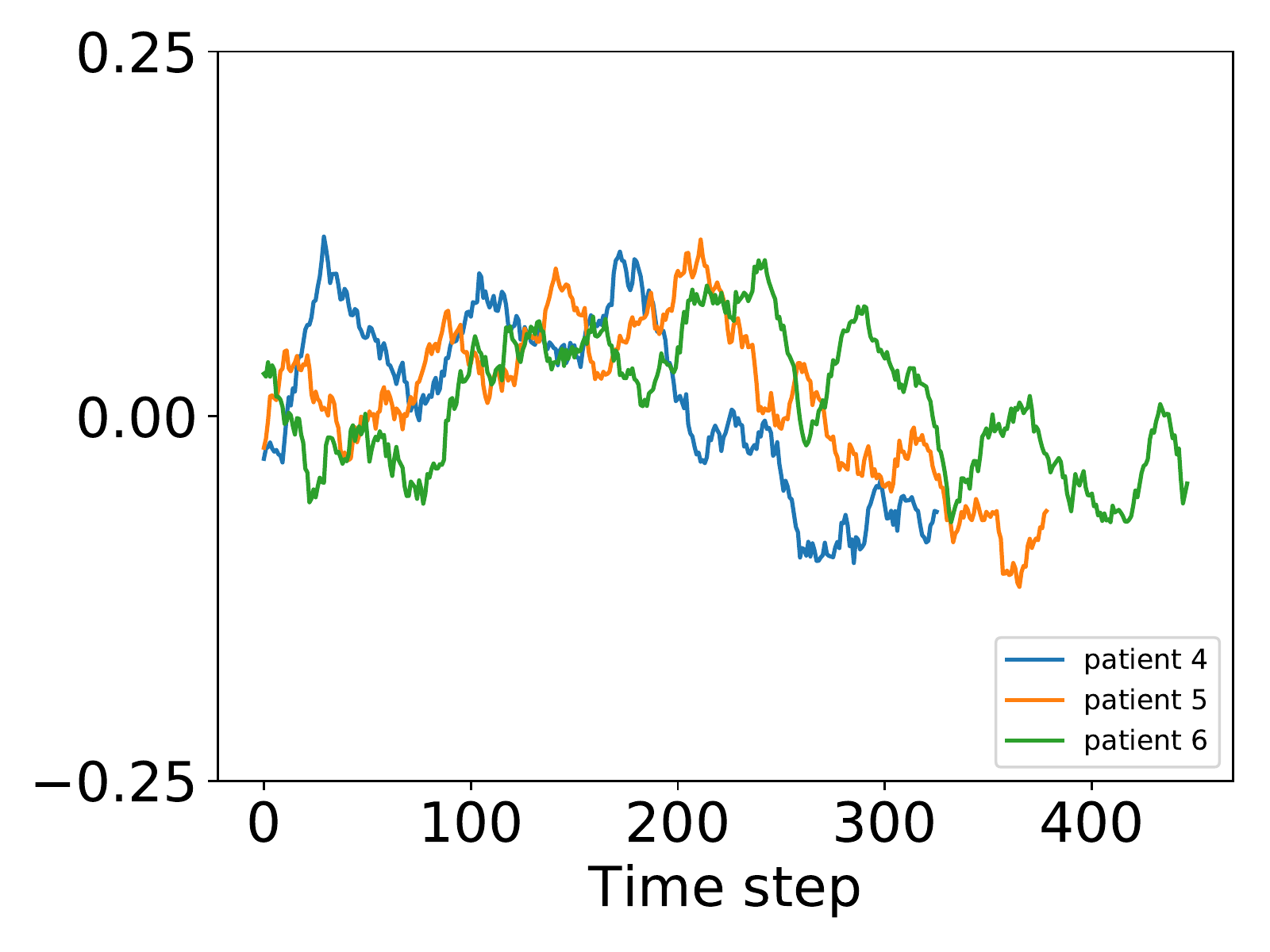}
\caption{IFN$\gamma$}
\label{fig:actions-6}
\end{subfigure}
\caption{Moving average (window length 20) of action values for PAF, IL-1, and IFN$\gamma$, selected by the learned policy during treatment of three patient parameterizations. (a)-(c): Varying recurrent injury parameter, with all other parameters held constant. (d)-(f): Varying initial injury size, with all other parameters held constant, and with recurrent injury set to zero. All patients healed from the policy's intervention.}
\label{fig:actions-timeseries}
\end{figure}

To investigate whether the learned policy indeed (1) is adaptive to patients' state progression over time, (2) prescribes personalized (i.e. patient-specific) actions, and (3) involves mediating multiple cytokines simultaneously in coordination, we characterized the policy by observing the time series of selected actions over various characteristic episodes on different patient parameterizations. Note that the five external parameters that define a patient parameterization are not provided to the RL algorithm in training nor when executing the policy in testing. Therefore, any variation in action choices over time and across different patients is due solely to observing the 21 systemic measurements during the course of treatment. 

In the first test, we selected patient parameterizations with recurrent injury values in the set $\lbrace 1, 5, 9 \rbrace$ (low, middle, and high values in the parameter range), and identical values for all other parameters. These parameterizations correspond to baseline mortalities $\lbrace 64\%, 96\%, 96\% \rbrace$, respectively. Each parameterization healed with the policy's intervention. \Cref{fig:actions-1,fig:actions-2,fig:actions-3} illustrate the moving average of time series of actions prescribed for cytokines PAF, IL-1, and IFN$\gamma$. The policy's actions for this patient demonstrate that the policy is adaptive, even switching from down-regulation to up-regulation during the course of treatment in the case of PAF. Differences between patient curves show the policy's specificity to different patient parameterizations. The periodicity of action values that arise after certain time points further reflects the policy's adaptive response to recurrent injuries, after patients have reached the health threshold (i.e. the policy results in a repeating cycle of up-regulation of IL-1 when the recurrent injury re-appears, and down-regulation once the recurrent injury is cleared). In each of the cytokines shown, we see that Patient 3 (green) takes the most time to enter the periodic region, since this patient has the highest recurrent injury number.

We repeated this experiment for patient parameterizations with initial injury size in the set $\lbrace 20, 24, 30 \rbrace$, with all other parameters held constant and recurrent injury set to zero (to remove the aforementioned periodicity). These parameterizations correspond to baseline mortalities $\lbrace 4\%, 30\%, 100\% \rbrace$, respectively. \Cref{fig:actions-4,fig:actions-5,fig:actions-6} demonstrate the policy's adaptive transition between up-regulation and inhibition, as well as dependence on the severity of patients' initial injury.

While a thorough biomedical interpretation of the learned policy is beyond the scope of this article, we note that the policy has some intuitive characteristics. For example, the control pattern of IL-1 (a pro-inflammatory cytokine) seen in \Cref{fig:actions-5} is consistent with non-suppression during the very early phases of the response where IL-1 is needed to clear the infection, followed by later suppression of IL-1 to aid in containment of runaway inflammation and collateral damage. Moreover, the suppression occurs later for patients with larger initial injury (e.g. Patient 6), likely due to the fact that the immune system requires a longer period of inflammation for larger initial injury sizes to contain and control the greater level of infection/injury present. Additionally, the pattern of IFN$\gamma$ control (\Cref{fig:actions-6}), also a pro-inflammatory cytokine but one that has additional effects modulating T\textsubscript{h} cell population distributions, could be interpreted as qualitatively following a similar pattern, but less clearly defined given its more complicated effects. The challenge and limitations of such post-hoc interpretation of the data might be addressed in the future with more in-depth representation of the various action spaces that could provide either substantiation of an interpretation, or, conversely, lead to additional non-intuitive insights into what underlying functions are important in the system's response and recovery.


\section*{Conclusion}

We investigated whether adaptive, personalized multi-cytokine mediation is an effective approach for lowering patient mortality for sepsis. We used an agent-based simulation of sepsis to train a treatment policy in which systemic patient measurements are used in a feedback loop to inform future treatments. Using deep reinforcement learning, we identified a policy that achieves 0\% mortality on the patient parameterization on which it was trained. More importantly, this policy also achieves 0.8\% mortality over 500 randomly selected patient parameterizations with baseline mortalities ranging from $1 - 99\%$ (with an average of 49\%) spanning the entire clinically plausible parameter space of the simulation. These results suggest that adaptive, personalized multi-cytokine mediation therapy could be a promising approach for treating sepsis. We hope that this work motivates researchers to consider such adaptive approaches as part of future clinical trials. 

In future work, we plan to explore the bounds of controllability, especially as they pertain to clinically relevant and tractable solutions. For example, we will seek effective policies that rely on less frequent patient measurements, apply actions less often, and only mediate cytokines for which existing drugs are known. Such a study could suggest a set of ``requirements'' for an adaptive, personalized multi-cytokine meditation therapy strategy to be effective.

To the best of our knowledge, this work is the first to consider adaptive, personalized multi-cytokine mediation therapy for sepsis, and is the first to exploit deep reinforcement learning to control a biological simulation. This work has demonstrated that mechanism-based simulations in the biomedical sciences can be a powerful alternative to relying exclusively on real-world data sets. Indeed, the mechanism-based simulation enabled us to explore therapeutic strategies that have never been attempted in clinical or experimental settings. This work has also demonstrated that deep reinforcement learning is a powerful methodology to control simulations of real-world systems; we hope that such methods continue to be applied to simulations in biology and beyond.

\subsection{Acknowledgments}
We thank Thomas Desautels, Jason Lenderman, and Aaron Wilson for their insightful feedback and comments throughout the course of this study. This work was supported in part by NIH/NIGMS Grant 1S10OD018495-01 (G.A. and C.C.).

\subsection{Author contributions}
B.P., C.C., and G.A. developed the IIRABM. B.P. designed and implemented the sepsis environment. B.P., J.Y., and W.G. implemented DDPG. B.P. and C.S. integrated the IIRABM with Python. B.P., J.Y., and W.G. executed the experiments. B.P., J.Y., C.C., and D.F. analyzed the experimental results. B.P., J.Y., and D.F. prepared the figures and wrote the manuscript. G.A. provided biological and simulation expertise and guidance. D.F. conceived of and directed the project. All authors provided discussions, commentary, and manuscript feedback.

\subsection{Competing interests}
The authors declare no competing financial interests.

\bibliography{sample}

\end{document}